\def\year{2018}\relax
\documentclass[letterpaper]{article} 
\usepackage{aaai18}  
\usepackage{times}  
\usepackage{helvet}  
\usepackage{courier}  
\usepackage{url}  
\usepackage{graphicx}  
\usepackage{amsfonts}       
\usepackage{multirow}         
\usepackage{multicol} 
\usepackage{amsmath}
\usepackage{rotating}
\usepackage{multirow}
\usepackage{comment}
\usepackage{amsthm}
\usepackage[ruled]{algorithm2e}
\usepackage{multicol}

\newtheorem{proposition}{Proposition}
\newcommand*\diff{\ d}

\newcommand\backtiny{\hspace{-0.1em}}
\newcommand\backq{\hspace{-0.25em}}
\newcommand\backs{\hspace{-0.6em}}
\newcommand\backm{\hspace{-1.1em}}

\newcommand \pp{P}

\begin{document}
%
\title{Multiagent Soft Q-Learning}
\author{Ermo Wei \and Drew Wicke \and David Freelan \and Sean Luke\\
Department of Computer Science, George Mason University, Fairfax, VA USA\\
ewei@cs.gmu.edu,\ \ dwicke@gmu.edu,\ \ dfreelan@gmu.edu,\ \ sean@cs.gmu.edu
}
\maketitle
\begin{abstract}
Policy gradient methods are often applied to reinforcement learning in continuous multiagent games.  These methods perform local search in the joint-action space, and as we show, they are susceptable to a game-theoretic pathology known as {\it relative overgeneralization}.  To resolve this issue, we propose Multiagent Soft Q-learning, which can be seen as the  analogue of applying Q-learning to continuous controls.  We compare our method to MADDPG, a state-of-the-art approach, and show that our method achieves better coordination in multiagent cooperative tasks, converging to better local optima in the joint action space.
\end{abstract}

\section{Introduction}
Multiagent reinforcement learning (or MARL) is a type of Reinforcement Learning (RL) involving two or more agents.  The mechanism is similar to traditional reinforcement learning: the environment is some current {\it state} (which the agent can only sense through its observation), the agents each perform some {\it action} while in that state, the agents each receive some {\it reward}, the state transitions to some new state, and than the process repeats.  However in MARL, both transitions from state to state and the rewards allotted are functions of the {\it joint action} of the agents while in that state.  Each agent ultimately tries to learn a {\it policy} that maps its observation to the optimal action in that state: but these are individual actions, not joint actions, as ultimately an agent cannot dictate the other agents' actions.

Multiagent Learning has been investigated comprehensively in discrete action domains. Many methods have been proposed for {\it  equilibrium learning}~\cite{Littman1994Markov,Littman2001Friend,Hu2003Nash,Greenwald2003Correlated}, where the agents are trying to learn policies that satisfy some equilibrium concept from Game Theory. Almost all the equilibrium learning methods that have been proposed are based on off-policy Q-learning. This is not surprising, as multiagent equilibrium learning is naturally off-policy, that is, the agents are trying to learn an equilibrium policy while exploring the environment by following another policy. However, this situation does not apply to continuous games, that is, games with continuous actions. When RL must be applied to continuous control, policy gradient methods are often taken into consideration. However, in the past, it does not combine with off-policy samples as easily as the tabular Q-Learning. For this reason, RL has not been able to achieve as good performance in continuous games as it has in discrete domains.

In this paper, we consider cooperative games, where the agents all have the same reward function. Cooperative MARL problems can be categorized based on how much information each agent knows. If we have a central controller to control the learning process of each agent, then we have centralized training with decentralized execution~\cite{Oliehoek2008Optimal}. If the agents are learning concurrently, and each agent is told what the other agent or agents did, then the problem is known as a {\it joint action learner} problem. If the agents are learning concurrently but are not told what the others did, then we have an {\it independent learner} problem.  

\begin{figure}[h]
\begin{center}
\includegraphics[width=0.4\textwidth]{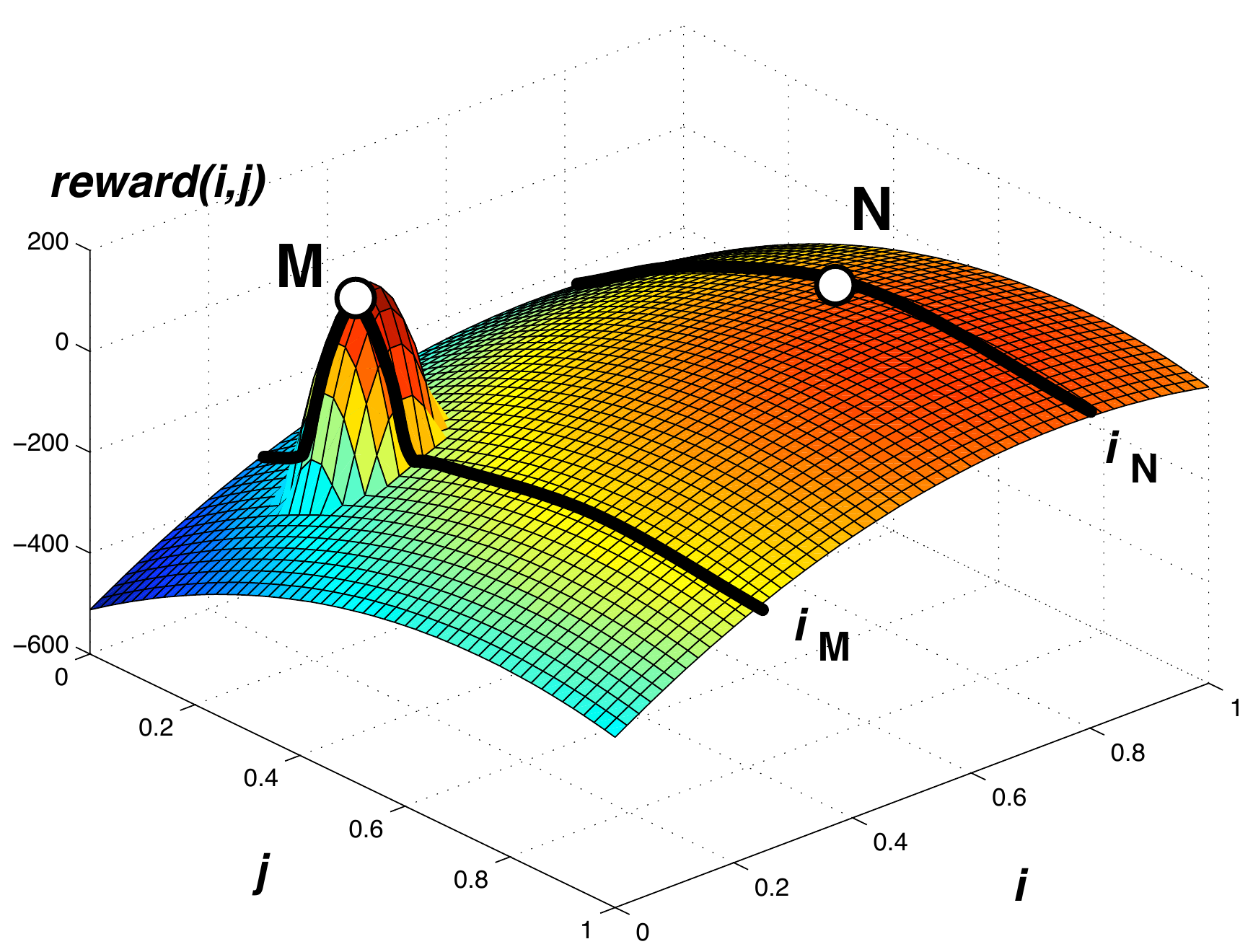}
\end{center}
\caption{The relative overgeneralization pathology in continuous games. 
}
\label{fig:joint}
\end{figure}

When the information is limited for learners in cooperative games, as is the case with independent learners, a pathology called {\it relative overgeneralization} can arise~\cite{Wei2016Lenient}. Relative overgeneralization occurs when a suboptimal Nash Equilibrium in the joint space of actions is preferred over an optimal Nash Equilibrium because each agent's action in the suboptimal equilibrium is a better choice when matched with arbitrary actions from the collaborating agents. For instance, consider a continuous game in Figure~\ref{fig:joint}. The axes \(i\) and \(j\) are the various actions that agents \(A_i\) and \(A_j\) may perform (we assume the agents are performing deterministic actions), and the axis \(\text{rewards }(i,j)\) is the joint reward received by the agents from a given joint action \(\langle i,j\rangle\). Joint action \(M\) has a higher reward than joint action \(N\).  However, the average of all possible rewards for action \(i_M\), of agent \(A_i\) is lower than the average of all possible rewards for action \(i_N\). Thus, the agents tend to converge to N.


In this paper, we first analytically show how relative overgeneralization prevents policy gradient methods from achieving better coordination in cooperative continuous games. This is even true in centralized training if we are not using the information wisely. Then we tackle the relative overgeneralization problem in these games by introducing Multiagent Soft Q-Learning, a novel method based on Soft Q-Learning and deep energy-based policies~\cite{Haarnoja2017Reinforcement}. Our method is similar to MADDPG~\cite{Lowe2017Multi}, a recently proposed centralized learning algorithm. Thus, it belongs to the centralized training with decentralized execution paradigm. In this setting, since the training is centralized and we use the information wisely, it avoids the co-adaptation problem in Multiagent RL and greatly reduces the sample complexity, as the environment for agents is stationary.
\section{Background}
In this section, we first give an introduction to Markov Decision Processes (MDP) and various generalizations. We then introduce policy gradient methods. 
\subsection{Markov Decision Processes and Stochastic Games}
A Markov Decision Process (or MDP) can be used to model the interaction an agent has with its environment.  An MDP is a tuple \(\{S, A, T, R, \gamma, H\} \) where  \(S\) is the set of states; \(A\) is the set of actions available to the agent; \(T\) is the transition function \(T(s, a, s') = {\pp}(s'|s,a)\) defining the probability of transitioning to state \(s'\in S\) when in state \(s\in S\) and taking action \(a\in A\); \(R\) is the reward function \(R:S \times A \mapsto \mathbb{R}\); \(0 < \gamma < 1\) is a discount factor; and \(H\) is the horizon time of the MDP, that is, the number of steps the MDP runs.\footnote{Any infinite horizon MDP with discounted rewards can be \(\epsilon\)-approximated by a finite horizon MDP using a horizon 
\(H_\epsilon = \frac{\log_{\gamma}(\epsilon(1-\gamma))}{\max_{s,a} | R(s,a)|}\)~
\cite{Jie2010Connection}}  An agent selects its actions based on the policy \(\pi_\theta(a|s)\), which is a distribution over all possible actions \(a\) in state \(s\) parameterized by \(\theta \in \mathbb{R}^n\). 

The concept of an MDP can be extended to partially observable (POMDP) settings, where agents do not directly sense the state \(s\). Rather, they receive some observation \(o\) sampled from a distribution conditioned on \(s\).

MDPs can also be generalized to a cooperative multiagent settings, called a {\it Cooperative Stochastic Game} or CSG. This is a game with \(n\) agents (or players),  defined by the tuple \(\{S,\mathcal{A},R, T, \gamma, H\}\), where \(S\) is the state space, \(\mathcal{A}=A^{1}\times ... \times A^{n}\) is the joint action space of \(n\) agents, \(R:S \times \mathcal{A} \rightarrow \mathbb{R}\) is the reward function for each agent \(i\), and \(T(s,\vec{a},s') = {\pp}(s'|s,\vec{a})\) is the transition function, where \(\vec{a} = \langle a^1 \cdots a^n\rangle \in \mathcal{A}\) is the joint action of all agents. Thus the reward the agents receive and the state to which they transition depends on the current state and agents' joint action.  Each agent \(i\) determines its action using a policy \(\pi^i\).   We will also use \(-i\) to denote all agents except for agent \(i\).  A POCSG can be thought as taking CSG into a partially observable setting.

In the multiagent setting, a rational agent will play its \textit{best response} to the other agents' strategy.  If all agents are following a policy that implements this strategy, they will arrive at a Nash equilibrium defined as a solution where \(\forall i\ R_i(s,\pi_1^*,\dots \pi_i^* \ldots\pi_n^*)\geq R_i(s,\pi_1^*,\ldots,\pi_{i-1}^*,\pi_i,\pi_{i+1}^*,\ldots,\pi^*_n)\) for all of the strategies \(\pi_i\) available to agent \(i\).  \(\pi_i^*\) denotes the best response policy of agent \(i\). 

\subsection{Policy Gradient Methods}
In single agent continuous control tasks, it is common to apply a {\it policy gradient} method to determine an optimal policy. We describe that process here.  To start, we define the expected return \(J(\theta)\) of a policy \(\pi_{\theta}\) as
\begin{equation}
J(\theta) =  E_{{\pp}_{\theta}(\tau)}\big[R(\tau)\big] = \sum_{\tau} P_{\theta}(\tau)R(\tau),
\label{eq:base}
\end{equation}

\noindent where \(P_{\theta}(\tau)\) is the probability distribution over all possible state-action trajectories \(\tau = \langle s_0, a_0, s_1, a_1, \ldots, s_H, a_H\rangle\) induced by following policy \(\pi_{\theta}\), and \(R(\tau) = \sum_{t=0}^{H}\gamma^{t} R(s_t,a_t)\) is the discounted accumulated reward along trajectory \(\tau\). We want to compute the gradient \(\nabla_\theta J(\theta)\), so that we can follow the gradient to a local optimum in the space of policy parameters.  To do this we use the likelihood-ratio trick~\cite{Williams1992Simple}, where we write the gradient as
\begin{equation}
\begin{split}
	\nabla_{\theta} J(\theta) = &\sum_{\tau} \nabla_{\theta} P_{\theta}(\tau)R(\tau) = \sum_{\tau} P_{\theta}(\tau) \frac{\nabla_{\theta} P_{\theta}(\tau)}{P_{\theta}(\tau)}R(\tau) \\
	=& E_{{\pp}_{\theta}(\tau)}\big[\nabla_{\theta} \ln P_{\theta}(\tau) R(\tau)\big]
	\label{eq:gradient}
\end{split}
\end{equation}

\noindent and estimate it by performing \(m\) sample trajectories \(\langle\tau^{(1)},\ldots,\tau^{(m)}\rangle\), calculating the corresponding terms, and then taking the average, that is, \(\nabla_{\theta} J(\theta) \approx \frac{1}{m}\sum^{m}_{j=1} \nabla_{\theta} \ln P_{\theta}(\tau^{(j)}) R(\tau^{(j)})\). This policy gradient method can also use off-policy samples by introducing importance sampling, where we scale each term in the empirical expectation by \(\frac{P_{\theta}(\tau)}{Q(\tau)}\), where \(Q\) is another distribution from which our off-policy samples come. The intuition behind Equation~\ref{eq:gradient} is that the reward term \(R(\tau)\) scales the gradient proportionally to the reward along the trajectories. 

One problem with using this likelihood-ratio estimator in practice is that it suffers from a large variance, and thus requires a great many samples to give an accurate estimation. There are various methods proposed to deal with this. A first approach is to replace the Monte Carlo estimation of the reward along trajectories \(R(\tau)\) with a value function.  This leads to the {\it Stochastic} (SPG) and {\it Deterministic Policy Gradient }(DPG) Theorems~\cite{Sutton1999Policy,Silver2014Deterministic}, shown below respectively:
\[
\begin{split}
\nabla_{\theta}J(\theta) & = \int_S \backq \rho^{\pi_{\theta}}(s)\int_A \backq \nabla_{\theta} \pi_{\theta}(a|s) Q^{\pi_{\theta}}(s,a) \diff a\diff s \\
&= E_{s \sim \rho^{\pi_{\theta}}, a \sim \pi_{\theta}} \bigg[\nabla_{\theta} \ln \pi_{\theta}(a|s) Q^{\pi_{\theta}}(s,a)\bigg] \\
\nabla_{\theta}J(\theta) & = \int_S \backq  \rho^{\pi_{\theta}}(s) \nabla_{\theta} \pi_{\theta}(s) \nabla_a Q^{\pi_{\theta}}(s,a) |_{a = \pi_{\theta}(s)} \diff s\\
& = E_{s \sim \rho^{\pi_{\theta}}} \bigg[\nabla_{\theta} \pi_{\theta}(s) \nabla_a Q^{\pi_{\theta}}(s,a) |_{a = \pi_{\theta}(s)}\bigg],
\end{split}
\]
\noindent where \(\rho^{\pi_{\theta}}(s') = \int_{S} \sum_{t=1}^{\infty} \gamma^{t-1} {\pp}(s) {\pp}(s\rightarrow s', t , \pi_\theta)\diff s\) is the discounted distribution over states induced by policy \(\pi_\theta\) and starting from some state \(s \in S\). Specifically, \({\pp}(s\rightarrow s', t , \pi)\) is the probability of going \(t\) steps under policy \(\pi\) from state \(s\) and ending up in state \(s'\). The theorems introduced a class of algorithms~\cite{Peters2008Natural,Degris2012Linear} under the name \textit{actor-critic} methods, where the {\it actor} is the policy \(\pi\) and the {\it critic} is the Q-function. 

The actor-critic algorithms have also been used in an off-policy setting through importance sampling~\cite{Degris2012Linear}. Recently, another method called a {\it replay buffer}~\cite{Mnih2013Playing} has drawn people's attention for being able to do off-policy learning with actor-critic algorithms~\cite{Lillicrap2015Continuous}. In this method, we store all the samples in a buffer and at every step of learning we sample a mini-batch from this buffer to estimate the gradient of either Q-Function or policy.

\section{Related Work}
The idea of learning to cooperate through policy gradient methods has been around for a long time, but mainly for discrete action domains~\cite{Banerjee2003Adaptive}. ~\citeauthor{Peshkin2000Learning} have applied the REINFORCE policy gradient to both CSG and POCSG tasks. However, as we will show later, a naive use of this gradient estimator is dangerous in the multiagent case. ~\citeauthor{Nair2003Taming} proposed {\it Joint Equilibrium-Based Search for Policies} (JESP), applied to POCSGs. The main idea here is to perform policy search in one agent while fixing the policies of other agents. Although this method is guaranteed to converge to a local Nash Equilibrium, it is essentially a round-robin single agent algorithm.

Recently, with the boom of Deep Reinforcement Learning (DRL), deep MARL algorithms have been proposed to tackle large scale problems. One of the main streams is the centralized training with decentralized execution.~\citeauthor{Foerster2016Learning} proposed a method to learn communication protocols between the agents. They use inter-agent backpropagation and parameter sharing.~\citeauthor{Foerster2017Stabilising} studied how to stablize the training of multiagent deep reinforcment learning using importance sampling. Two actor-critic algorithms have been proposed in~\cite{Foerster2017Counterfactual,Lowe2017Multi}. They argue that by using a central critic we can ease the training of multiple agents, and that by keeping a separate policy, the agent can execute with only its local information, which makes it possible to learn in POCSGs. Among these two algorithms, MADDPG~\cite{Lowe2017Multi} is most relevant to us. It uses the learning rule from DDPG~\cite{Lillicrap2015Continuous} to learn a central off-policy critic based on Q-learning, and uses the following gradient estimator to learn the policies for each agent \(i\):
\[
\nabla_{\theta^i} J(\theta^i) = E_{s, a^{-i} \sim D}\bigg[\nabla_{\theta^i}\pi^{i}_{\theta_i}(a_i|o_i)\nabla_{a_i}Q(s, \vec{a})|_{a_i = \pi^{i}_{\theta_i}(o_i)}\bigg],
\]
where \(\theta^i\) is the agent \(i\)'s policy parameters, \(D\) is the replay buffer, and \(o_i\) is the local observation of agent \(i\). During the centralized training process, the critic has access to the true state \(s = [o_1, \ldots, o_n]\). But at execution time, each agent only has access to \(o_i\).
\section{Multiagent Actor-Critic Algorithms}
As we described earlier, if we have limited information for our agent, we can suffer from the relative overgeneralization problem. In this section, we demonstrate how this affects the actor critic algorithm. We will first derive the policy gradient estimator for the cooperative multiagent case and then discuss several problems that occurs if we naively use this estimator, which will shed light on the reasons why Multiagent Soft Q-Learning may be useful.

\begin{proposition} 
\label{pro:multiagent}
For any episodic cooperative stochastic game with \(n\) agents, we have the following multiagent stochastic policy gradient theorem:
\begin{multline*}
\nabla_{\theta^i} J(\vec{\theta}) =  \int_{s} \rho^{\pi^1,\cdots, \pi^n}(s) \int_{A^i} \nabla_{\theta} \pi(a^i|s)\\ \int_{A^{-i}} \pi^{-i}(a^{-i}|s) Q^{\pi^1, \cdots, \pi^n}(s, \vec{a})\diff a^{-i}\diff a^{i} \diff s
\end{multline*}
\end{proposition}

The proof of this proposition is provided in Proof A at the end of this paper.  From Proposition~\ref{pro:multiagent}, we can see that the policy gradient for agent \(i\) at each state is scaled by \(Q^{\pi^{i}}(s, a^i) = \int_{A_{-i}} \pi^{-i}(a^{-i}|s) Q^{\pi^1 \cdots \pi^n}(s, \vec{a}) da^{-i}\), which are the joint-action Q-values averaged by the other agents' policies. Their are several problems with this estimator. First, for any agent \(i\), the joint-action Q-function is an on-policy Q-function. That is, it is learned under policy \(\pi^i\), and \(\pi^{-i}\) which is not the best response policy of other agents. Thus, the joint-action Q-function may not scale the gradient in right magnitude. Second, if we are an independent learner and play as \(A_i\) in game shown in Figure~\ref{fig:joint}, we only have access to \(Q^{\pi^{i}}(s, a^i)\), since this value is averaged by other's policies, the value of the action \(\langle i_N\rangle\) would be higher than the value of action \(\langle i_M\rangle\) even under the optimal Q-function, thus mistakenly scaling the gradient to towards \(\langle i_N\rangle\). 

MADDPG solves the previous two issues by using the following methods. First, it uses the replay buffer~\cite{Lillicrap2015Continuous} to learn an off-policy optimal Q-function very much like what we learn for Q-Learning~\cite{Silver2014Deterministic}. This is not doable with traditional importance sampling based off-policy learning. Second, it's using the centralized training method which gives it direct access to the joint-action Q-function, but not the policies. 

However, MADDPG fails to use the optimal action for gradient scaling, making it still vulnerable to the relative overgeneralization problem. To see that, consider its gradient estimator,
\[
\nabla_{\theta^i} J(\theta^i) = E_{s, a^{-i} \sim D}\bigg[\nabla_{\theta^i}\pi^{i}_{\theta_i}(a_i|o_i)\nabla_{a_i}Q^{*}(s, \vec{a})|_{a_i = \pi^{i}_{\theta_i}(o_i)}\bigg]
\]
We see that for agent \(i\), it tries to ascend the policy gradient based on \(Q^{*}(s, \vec{a})\), where \(a^{-i}\) is from the replay buffer \(D\) rather than the optimal policy, which is another way of averaging the Q-values based on others' policies. As we showed in Figure~\ref{fig:joint}, this average-based estimation can lead to relative overgeneralization.

\section{Multiagent Soft Q-Learning}
In this paper, we propose MARL method for cooperative continuous games. We show that on the one hand, our method is an actor-critic method, which thus can benefit from the centralized training method, with one central critic and multiple distributed policies. And on the other hand, our method resembles Q-Learning, and thus, it efficiently avoids the relative overgeneralization problem. We first introduce Soft Q-Learning and then describe how we use it for multiagent training.

\subsection{Soft Q-Learning}
Although Q-Learning has been widely used to deal with control tasks, it has many drawbacks. One of the problems is that at the early stage of learning, the \(\max\) operator can bring bias into the Q-value estimation~\cite{Fox2016Taming}. To remedy this,  Maximum Entropy Reinforcment Learning (MERL) was introduced, in which tries to find the following policy:
\[
\pi^{*}_{\text{MaxEnt}} = \text{argmax}_{\pi} \sum_{t} E_{(s_t, a_t) \sim \rho^{\pi}}[r(s_t,a_t) + \alpha \mathcal{H}(\pi(\cdot|s)]
\]
where \(\mathcal{H}(\pi(\cdot|s))\) is the entropy of the policy. The parameter \(\alpha\) controls the relative importance of the reward and entropy: when it goes to 0, we recover ordinary RL. From this objective, a learning method similar to Q-Learning can be derived, called Soft Q-Learning~\cite{Haarnoja2017Reinforcement}. Its learning algorithm is
\[
\begin{split}
Q_{\text{soft}}(s_t,a_t) &\leftarrow r_t + \gamma E_{s_{t+1}}[V_{\text{soft}}(s_{t+1})] \:\:\:\forall s_t, a_t,\\
V_{\text{soft}}(s_{t}) &\leftarrow \alpha \log \int_A \exp(\frac{1}{\alpha} Q_{\text{soft}}(s_t,a')) da'.
\end{split}
\]
\citeauthor{Haarnoja2017Reinforcement} have shown that by using this update rule, \(Q_\text{soft}\) and \(V_\text{soft}\) can converge to \(Q^{*}_\text{soft}\) and \(V^{*}_\text{soft}\) respectively, and by driving \(\alpha \rightarrow 0\), Q-learning with a hard max operator can be recovered. For this reason,~\citeauthor{Haarnoja2017Reinforcement} named this Soft Q-learning. 

Once we have the learned Q-function above, we can get the optimal max entropy policy as
\[
\pi^{*}_{\text{MaxEnt}}(a_t|s_t) = \exp(\frac{1}{\alpha}Q^{*}_{\text{soft}}(s_t, a_t) - V^{*}_{\text{soft}}(s_t)) \propto Q^{*}_{\text{soft}}(s_t, a_t).
\]
 A nice property of this policy is that it spreads widely over the entire action space in continuous control tasks. Thus, if we have such a policy, and if there are multiple modes in the action space, we can find them much more effectively than with more deterministic policies (e.g. Gaussian policy) which are typically used in actor-critic algorithms. However, since the form of this policy is so general, sampling from it is very hard. Soft Q-Learning solves this issue by using Stein Variational Gradient Descent (SVGD)~\cite{Liu2016Stein} to approximate the optimal policy through minimizing the KL-divergence:
 \begin{equation}
 D_{\text{KL}} = \bigg( \pi_{\theta}(\cdot|s_t)\big|\big|\exp\big(\frac{1}{\alpha}Q^{*}_{\text{soft}}(s_t, a_t) - V^{*}_{\text{soft}}(s_t)\big)\bigg),
 \label{eq:kl}
 \end{equation}
 where policy \(\pi_{\theta}(\cdot|s)\) is our approximate policy. Since \(-\frac{1}{\alpha}Q^{*}_{\text{soft}}(s_t, a_t)\) can be viewed as  an energy function, and the authors are using a deep neural network to approximate the Q-function, they call this a {\it deep energy-based} policy.
 It has been demonstrated that using the Soft Q-Learning with deep energy based policies can learn multimodal objectives. In Soft Q-Learning we need to learn both the Q-function and the energy-based policy \(\pi(\cdot|s)\). Thus, Soft Q-Learning can be thought as an actor-critic algorithm.
Now consider the multiagent case. To make it clear, we first recall how we can achieve coordination in a discrete domain. In discrete domains, when we have the \(Q^{*}(s,a)\) function, we simply apply the \textit{argmax} operator to it and then let each agent do its own part of the optimal action. This is possible since we can do global search in the joint-action space for a given state. Now, with Soft Q-Learning and a deep energy-based model, we can mimic what we did in the discrete case. In this situation, we start with a high \(\alpha\) to do global search in the joint-action space, then quickly anneal the \(\alpha\) to lock on some optimal action, like the argmax operator. It has been shown that by annealing the \(\alpha\), we can get a deterministic policy from deep energy-based policies~\cite{Haarnoja2017Reinforcement}.

\begin{algorithm}[h]
\SetAlgoLined
	\SetKwInOut{Input}{input}
	\Input{A central critic Q, N policies for all N agents, \(\alpha\), and the epoch start to annealing \(t\)}
	\For{episode = 0 to M}{
		Update central critic Q using the method from Soft Q-Learning.\\
		\For{agent = 1 to N}{
			Update the joint policy for agent \(i\) using equation (\ref{eq:kl})
		}
		\uIf{episode \(\ge\) t}{
			anneal \(\alpha\)
		}
	}
 \caption{Multiagent Soft Q-Learning}
 \label{algo:masoftq}
 \end{algorithm}

As we described before, Soft Q-Learning is also an actor-critic method. Thus, we can borrow the idea of learning a centralized joint action critic with Soft Q-Learning from MADDPG. Then for each of the agents, instead of learning  a mapping for its own observation to its own action, we learn a mapping from its own observation to the joint-action. When the agent interacts with the environment, it just performs its own part of the joint action. We start the learning with high \(\alpha\) value and let it explore the joint action space, we then quickly anneal the \(\alpha\) to let each agent find a better local optima in joint-action space. Our algorithm is given at Algorithm~\ref{algo:masoftq}.


\begin{figure}[h]
\begin{center}
\includegraphics[width=0.4\textwidth]{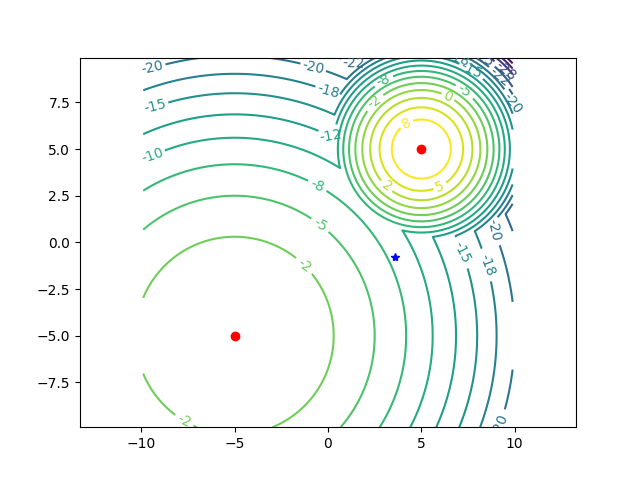}
\end{center}
\caption{The Max of Two Quadratic game. The dots mark the two local optima in the joint action space while the star marks the joint action of the two agents. The contour shows the reward level.}
\label{fig:two_quadratic}
\end{figure}

\begin{figure}[t]
\begin{center}
\includegraphics[width=0.5\textwidth]{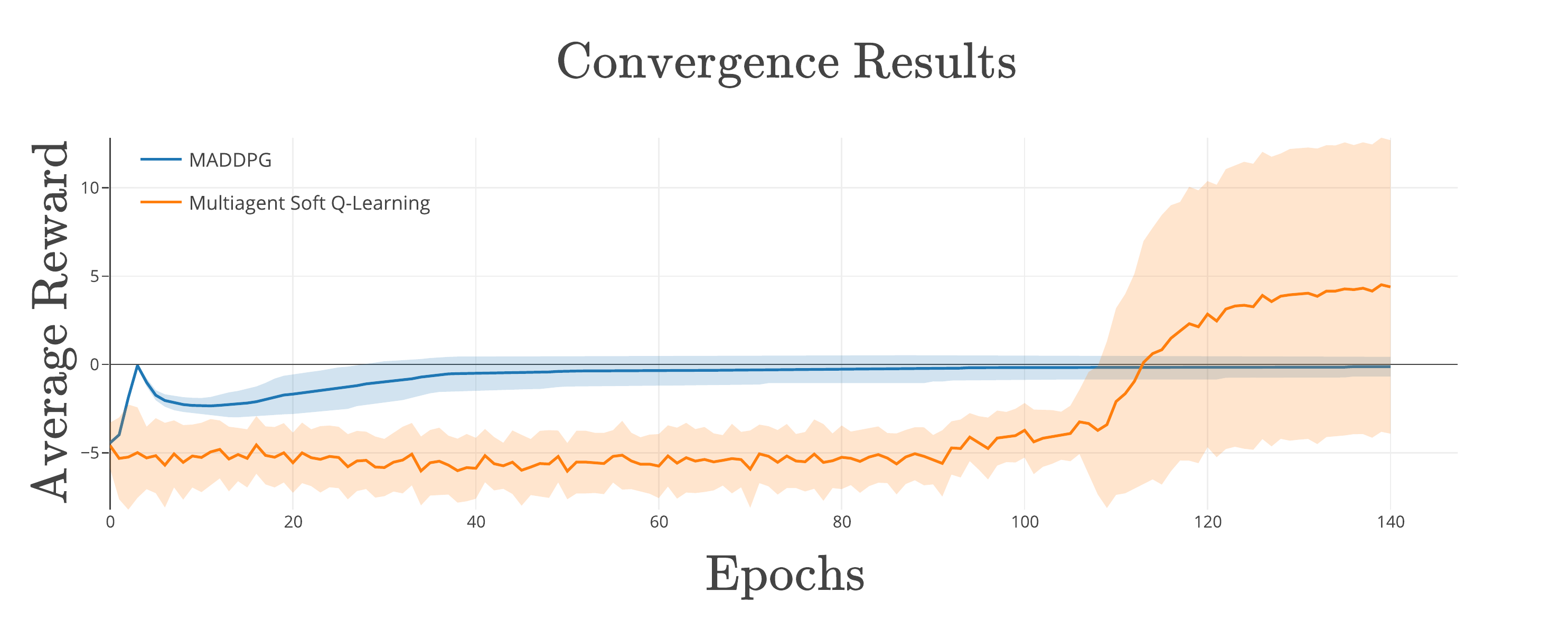}
\end{center}
\caption{The average reward for both algorithms. Multiagent Soft Q-Learning finds the better local optima quickly after we anneal \(\alpha\).}
\label{fig:average_reward}
\end{figure}

\section{Experiments}
To show that our Multaigent Soft Q-Learning method can achieve better coordination, we consider the Max of Two Quadractics game from previous literature~\cite{Panait2006Biasing}. This is a simple single state continuous game for 2 agents, one action dimension per agent. Each agent has a bounded action space. The reward for a joint action is given by following equation
\[
\begin{split}
&f_1 = h_1 \times \big[-\big(\frac{a_1 - x_1}{s_1}\big)^2 - \big(\frac{a_2 - y_1}{s_1}\big)^2\big]\\
&f_2 = h_2 \times \big[-\big(\frac{a_1 - x_2}{s_2}\big)^2 - \big(\frac{a_2 - y_2}{s_2}\big)^2)\big] + c\\
&r(a_1, a_2) = \max(f_1, f_2)
\end{split}
\]
where \(a_1, a_2\) are the actions from agent 1 and agent 2 respectively. In the equation above, \(h_1 = 0.8, h_2 = 1, s_1 = 3, s_2 = 1, x_1 = 5, x_2 = 5, y_1 = -5, y_2= - 5, c = 10\) are the coefficients to determine the reward surface (see Figure~\ref{fig:two_quadratic}). Although the formulation of the game is rather simple, it poses a great difficulty to gradient-based algorithms as, over almost all the joint-action space, the gradient points to the sub-optimal solution located at (-5, -5).

We trained the MADDPG agent along with our Multagent Soft Q-Learning agent in this domain. As this was a simple domain, we used two-hidden-layer networks with size \{100, 100\}, and we trained the agents for 150 epochs for 100 steps per epoch. The training was not started until we had 1000 samples in the replay buffer. Both agents scaled their reward by 0.1. For our Multiagent Soft Q-Learning agent, we started the annealing at epoch 100, and finished the annealing in 15 epochs. We started with \(\alpha = 1\), and annealed it to 0.001. For the rest of the parameters, we used the default setting from the original DDPG~\cite{Lillicrap2015Continuous} and Soft Q-Learning~\cite{Haarnoja2017Reinforcement} papers. In addition, to mimic the local observation setting where centralized learning was suitable, we gave the two agents in both algorithms different observation signals, where the first agent would always sense the state as \(\langle0\rangle\), and the second agent would always sense it as \(\langle1\rangle\). Then the state for the central critic was \(\langle 0, 1\rangle\).

The result is in Figure~\ref{fig:average_reward}. However, the plot is an average over all 50 experiment runs, and hence, may hide some critical information. On closer investigation, we found that Multiagent Soft Q-Learning converged to the better equilibrium {\bf 72\%} of the time, while MADDPG {\bf never} converged to~it. 


\section{Conclusion and Future Work}
In this paper, we investigated how to achieve better controls in continuous games. We showed why the traditional policy gradient methods is not suitable for these tasks, and why the gradient-based method can fail to find better local optima in the joint-action space. We then proposed Multiagent Soft Q-Learning based on the centralized training and decentralized execution paradigm, and showed that, we can achieve much better coordination with higher probability. And since we are using centralized training, the co-adaption problem can be avoided, thus, making our method sample-efficient compared to independent learners. We argue that Multiagent Soft Q-Learning is a competitive RL learner for hard coordination problems. 

There are some issues that we haven't been able to investigate thoroughly in this work. First, so far we have only applied our learner in the single state games. To better understand the algorithm, we would like to try our algorithm on sequential continuous games with hard coordination problems. Second, as we show in the experiment, with Soft Q-Learning we are not able to converge the better equilibrium for 100\% of the time. In the future, we would like to investigate different annealing methods to improve the convergence rate.  Last, we notice that Multiagent Soft Q-Learning models the joint action of all the agents, and thus the dimension of the action can explode with more agents. To solve this issue, we will investigate how to apply Soft Q-Learning in the independent learner case, where the algorithm scales well.

\section{Acknowledgments}
The research in this paper was conducted with the support of research infrastructure developed under NSF grant 1727303. We thank Tuomas Haarnoja and Haoran Tang for their helpful comments on implementing Soft Q-Learning.
\section{Proof A}
We first denote \(\vec{\pi}\) as the joint-policy. This proof requires that ${\pp}(s), {\pp}(s'|s, \vec{a}),  \pi^{i}(a^{i}|s), \nabla_{\theta^i} \pi^{i}(a^{i}|s),\) and \( Q^{\vec{\pi}}(s, \vec{a})$ be continuous in all parameters and variables $s, s',\vec{a}$. This regularity condition implies that $V^{\vec{\pi}}(s)\) and \(\nabla_{\theta^i} V^{\vec{\pi}} (s)$ are continuous functions of $\theta$ and $s$. \(S\) is also required to be compact, and so for any $\theta$, $\lvert\lvert  \nabla_{\theta^i} V^{\vec{\pi}} (s) \rvert\rvert$ is a bounded function of $s$.  The proof mainly follows along the standard Stochastic Policy Gradient Theorem. We assume agent $i$ follows the policy $\pi^i(a^i|s)$ parameterized by $\theta^i$. We denote \(\pi^{-i}\) as the joint policy of all agents but agent \(i\), and \(a^{-i}\) as the joint action of all agents except agent \(i\). For notation simplicity, we denote:
\[
\begin{split}
& \int_{A^{-i}} \pi^{-i}(a^{-i}|s) f(a^{-i}) \diff a^{-i}\\
= & \int_{A^1}\backs \pi^1(a^1|s) \cdots \int_{A^{i-1}} \pi^{i-1}(a^{i-1}|s)\\
& \int_{A^{i+1}} \pi^{i+1}(a^{i+1}|s) \cdots \int_{A^n}\backs \pi^n(a^n|s) f(a^{-i}) \\
& \diff a^n \cdots \diff a^{i+1} \diff a^{i-1} \cdots \diff a^1
\end{split}
\]
Using this new notation the proof follows:
	\[
	\begin{split}
	\hspace{-1em}
	& \nabla_{\theta^i} V^{\vec{\pi}}(s)\\
	= & \nabla_{\theta^i} \int_{A^1}\backs \pi^1(a^1|s) \cdots \int_{A^n}\backs \pi^n(a^n|s) \:Q^{\vec{\pi}}(s, \vec{a}) \diff \vec{a}\\
	= & \int_{A^{-i}}\backm \pi^{-i}(a^{-i}|s) \nabla_{\theta^i} \int_{A^i}\backs \pi^{i}(a^{i}|s) Q^{\vec{\pi}}(s, \vec{a}) \diff a^i \diff a^{-i}\\
	= &  \int_{A^{-i}}\backm \pi^{-i}(a^{-i}|s) \int_{A^i} \bigg[\nabla_{\theta^i} \pi^{i}(a^{i}|s) Q^{\vec{\pi}}(s, \vec{a}) \\
	& + \pi^{i}(a^{i}|s) \nabla_{\theta^i} Q^{\vec{\pi}}(s, \vec{a})\bigg] \diff a^i \diff a^{-i}\\
	= &  \int_{A^{-i}}\backm \pi^{-i}(a^{-i}|s) \int_{A^i}\bigg[\nabla_{\theta^i} \pi^{i}(a^{i}|s) Q^{\vec{\pi}}(s, \vec{a}) \\
	& + \pi^{i}(a^{i}|s) \int_S \backq \gamma {\pp}(s'|s, \vec{a}) \nabla_{\theta^i} V^{\vec{\pi}} (s') \diff s'\bigg] \diff a^i \diff a^{-i}\\
	\end{split}
	\]
\noindent We used Leibniz integral rule to exchange order of derivative and integration using the regularity condition and expanding $Q^{\vec{\pi}}(s, \vec{a})$ above.  We use \(P^{\pi}(s'|s, t)\) as short for \(P(s\rightarrow s', t, \pi)\). Now we iterate the formula,
	\[
	\begin{split}
	= &  \int_{A^{-i}}\backm \pi^{-i}(a^{-i}|s) \int_{A^i}\bigg[\nabla_{\theta^i} \pi^{i}(a^{i}|s) Q^{\vec{\pi}}(s, \vec{a})
	\hspace{\fill}+ \pi^{i}(a^{i}|s) \\
	& \int_{S}\backq \gamma {\pp}(s'|s,\vec{a}) \Big[ \int_{A^{-i}}\backm \pi^{-i}(a^{-i}|s') \int_{A^i}\backq \Big(\nabla_{\theta^i} \pi^{i}(a^{i}|s') Q^{\vec{\pi}}(s', \vec{a})\\
	& + \pi^{i}(a^{i}|s') \int_S \backq \gamma {\pp}(s''|s', \vec{a}) \nabla_{\theta^i} V^{\vec{\pi}} (s'') \diff s''\Big) \diff a^i \diff a^{-i} \Big] \diff s' \bigg] \\
	& \diff a^i \diff a^{-i}\\
	= &  \int_{A^{-i}}\backm \pi^{-i}(a^{-i}|s) \backtiny\int_{A^i}\backs \nabla_{\theta^i} \pi^{i}(a^{i}|s) Q^{\vec{\pi}}(s, \vec{a})\backq\diff a^i \backq\diff a^{-i} \\
	& + \backq\int_{S}\backq \gamma {\pp}^{\vec{\pi}}(s'|s,1)  \backtiny\int_{A^{-i}}\backm \pi^{-i}(a^{-i}|s') \backtiny \\
	& \int_{A^i}\backs \nabla_{\theta^i} \pi^{i}(a^{i}|s') Q^{\vec{\pi}}(s', \vec{a}) \backq\diff a^i \backq\diff a^{-i} \backq\diff s'\\
	& + \int_{S}\backq \gamma {\pp}^{\vec{\pi}}(s'|s,1) \int_{S}\backq \gamma {\pp}^{\vec{\pi}}(s''|s',1) \nabla_{\theta^i} V^{\vec{\pi}} (s'') \diff s''\diff s'\\
	= &  \int_{S} \sum_{t=0}^{\infty} \gamma^t {\pp}^{\vec{\pi}}(s'|s,t) \\
	& \int_{A^{-i}}\backm \pi^{-i}(a^{-i}|s') \int_{A^i}\backs \nabla_{\theta^i} \pi^{i}(a^{i}|s') Q^{\vec{\pi}}(s', \vec{a}) \diff a^i \diff a^{-i} \diff s'\\
	= &  \int_{S} \sum_{t=0}^{\infty} \gamma^t {\pp}^{\vec{\pi}}(s'|s,t) \\
	& \int_{A^i}\backs \nabla_{\theta^i} \pi^{i}(a^{i}|s')  \int_{A^{-i}}\backm \pi^{-i}(a^{-i}|s') Q^{\vec{\pi}}(s', \vec{a}) \diff a^{-i} \diff a^{i} \diff s'\\
	\end{split}
	\]
	

\noindent In the final line we use Fubini's theorem and exchange the order of integration using the regularity condition so that $\lvert\lvert  \nabla_{\theta^i} V^{\vec{\pi}} (s) \rvert\rvert$ is bounded. We then take the expectation over the possible start states \(s\):
	\[
	\begin{split}
	 \hspace{-1em}\nabla_{\theta^i} J(\theta) 
	 = & \nabla_{\theta^i} \int_{S}\backq {\pp}(s) V^{\vec{\pi}}(s) \diff s
	 \quad
	 = \int_{S}\backq {\pp}(s) \nabla_{\theta^i} V^{\vec{\pi}}(s) \diff s \\
	= &  \int_{S} \int_{S} \sum_{t=0}^{\infty} \gamma^t {\pp}(s) P^{\vec{\pi}}(s'|s, t) \diff s \int_{A_i} \backs \nabla_{\theta^i} \pi^{i}(a^{i}|s') \\
	& \int_{A_{-i}} \backm \pi^{-i}(a^{-i}|s') Q^{\vec{\pi}}(s', \vec{a}) \diff a^{-i} \diff a^{i} \diff s'\\
	= & \int_{S}\backq \rho^{\vec{\pi}}(s')  \int_{A^i}\backs \nabla_{\theta^i} \pi^{i}(a^{i}|s') \\ &\int_{A^{-i}}\backm \pi^{-i}(a^{-i}|s') Q^{\vec{\pi}}(s', \vec{a}) \diff a^{-i} \diff a^{i} \diff s'\qquad\qquad\qquad\qquad\qed
	\end{split}
	\]
%
\bibliography{policy}
\bibliographystyle{aaai}

\end{document}